\pdfoutput=1

\documentclass[11pt]{article}

\usepackage[preprint]{acl}
\usepackage{underscore}
\usepackage{times}
\usepackage{latexsym}

\usepackage[T1]{fontenc}

\usepackage[utf8]{inputenc}

\usepackage{microtype}

\usepackage{inconsolata}

\usepackage{graphicx}
\usepackage{epstopdf}
\usepackage{epsfig}
\usepackage{microtype}
\usepackage{subfigure}
\usepackage{amsfonts}
\usepackage{amssymb}
\usepackage{amsmath}
\usepackage{booktabs}
\usepackage{cleveref}
\usepackage{tabularx}
\usepackage{multirow}
\usepackage{makecell}

\title{Predicting Depression in Screening Interviews from Interactive Multi-Theme Collaboration}

\author{
 \textbf{Xianbing Zhao\textsuperscript{1,2}},
 \textbf{Yiqing Lyu\textsuperscript{2}},
 \textbf{Di Wang\textsuperscript{1}},
 \textbf{Buzhou Tang\textsuperscript{2}},
\\
\\
 \textsuperscript{1}Xidian University,
 \textsuperscript{2}Harbin Institute of Technology, Shenzhen,
}

\begin{document}
\maketitle
\begin{abstract}
Automatic depression detection provides cues for early clinical intervention by clinicians. Clinical interviews for depression detection involve dialogues centered around multiple themes. Existing studies primarily design end-to-end neural network models to capture the hierarchical structure of clinical interview dialogues. However, these methods exhibit defects in modeling the thematic content of clinical interviews: 1) they fail to explicitly capture intra-theme and inter-theme correlation, and 2) they do not allow clinicians to intervene and focus on themes of interest. To address these issues, this paper introduces an interactive depression detection framework. This framework leverages in-context learning techniques to identify themes in clinical interviews and then models both intra-theme and inter-theme correlation. Additionally, it employs AI-driven feedback to simulate the interests of clinicians, enabling interactive adjustment of theme importance. PDIMC achieves absolute improvements of 35\% and 12\% compared to the state-of- the-art on the depression detection dataset DAIC-WOZ, which demonstrate the effectiveness of modeling theme correlation and incorporating interactive external feedback. 
\end{abstract}

\section{Introduction}

Depression stands as one of the primary factors affecting individuals' mental health \cite{wei2022multi}, exerting negative impacts on their work and daily life through its influence on cognitive processes and behavioral patterns. Statistics indicate a year-on-year increase in the prevalence of depressive symptoms within populations in both China and the United States \cite{rinaldi2020predicting}, a trend that continues to escalate with advancements in detection methods. However, it remains a challenging task as it requires accurately capturing the relationships across hundreds of turns in clinical interview dialogues and predicting the depressive state \cite{mm-2-mallol2019hierarchical,Yang2023MentaLLaMAIM,Yao2024Depression}. 
\begin{figure}
    \centering
    \includegraphics[width=\linewidth]{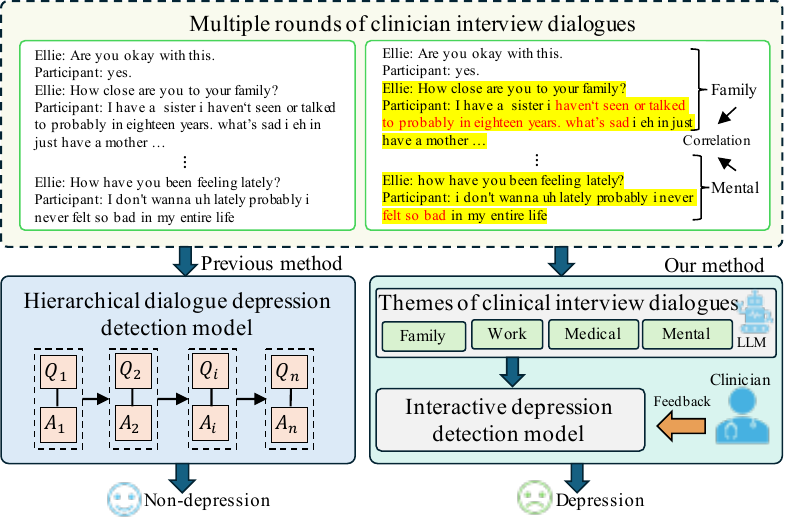}
    \caption{
Previous method only focused on modeling the sequential clinical interview dialogue information. Our method learns themes from clinical interview dialogues, models both intra-theme and inter-theme correlation, and finally introduces interactive feedback to guide the model in diagnosing depression.
    }
    \label{fig:premodel}
\end{figure}
Specifically, clinical interview dialogues are structured around multiple themes, such as family, work, and medical history. These themes exhibit inter-dependencies, and their correlations with depressive states vary in strength. As illustrated in Figure \ref{fig:premodel}, clinical interview dialogues begin with natural dialogues and then evolve around multiple themes \cite{daic-gratch2014distress,rinaldi2020predicting}, which are closely related to the depressive state of participant.

In recent years, significant efforts have been made to address the aforementioned challenges by modeling the hierarchical structure of clinical interview dialogues. Based on the type of hierarchical neural network employed, existing approaches can be broadly classified into two categories: 1) modeling the hierarchical dependencies in multi-turn question-answering \cite{wu2023self,anshul2023multimodal,zhang2024llms}. To facilitate the understanding of complex clinical interview dialogues, this approach focuses on capturing the correlation between individual question-answer pairs as well as the sequential dependencies across multiple turns. 2) capturing implicit themes \cite{gong2017topic,rinaldi2020predicting}. This approach relies solely on learnable parameters, modeling each question-answer pair's category as an implicit theme. To smoothly introduce themes, clinical interview dialogues often include dialogues that are unrelated to the depressive state. The former approach tends to trap the model in trivial details, while the latter fails to explicitly identify theme content. Moreover, neither of these methods allows clinicians to intervene in the model to focus on themes of interest.

To tackle these downsides, we introduce a novel interactive depression detection framework, namely \textbf{P}redicting \textbf{D}epression in Screening Interviews from \textbf{I}nteractive \textbf{M}ulti-Theme \textbf{C}ollaboration (PDIMC), which is the first depression detection framework incorporating explicit theme correlation learning and external feedback intervention. As illustrated in Figure \ref{fig:premodel}, clinical interview dialogues revolve around themes such as \textit{family}, \textit{work}, \textit{mental}, and \textit{medical}. To provide a more comprehensive evaluation, we additionally introduce a virtual theme \textit{overall}. Based on the above observations, we first design a theme-oriented in-context learning (TICL) module to extract theme content from complex clinical interview dialogues, preventing the model from being overwhelmed by trivial details. To capture both intra-theme and inter-theme correlation, we develop a theme correlation learning (TCL) module. Subsequently, we introduce a interactive theme adjustment strategy (ITAS), which leverages the large language model (LLM) to simulate clinician feedback, emphasizing key information from the feedback to further adjust the importance of different themes. Extensive experimental results on the well-known clinical interview depression dataset DAIC-WOZ demonstrate the effectiveness and superiority of our approach. Our main contributions are threefold:

\begin{itemize}
    \item We introduce an interactive depression detection framework for automated depression detection. To the best of our knowledge, this is the first framework that explores interactive depression detection using clinical interview data.
    \item We introduce a theme-oriented in-context learning technique to extract themes from clinical interview dialogues and design a theme correlation learning module to model both intra-theme and inter-theme correlation.
    \item  We propose an interactive theme adjustment strategy, which leverages the LLM to simulate clinician feedback, dynamically adjusting the importance of theme. This enables the model to focus on clinician-preferred themes for more effective depression detection.
\end{itemize}

\section{Related Work}

\subsection{In-context Learning}
In-context Learning (ICL) \cite{gpt-3-brown2020language,icl0-dong2022survey} is a technique that uses few-shot in-context learning samples to guide the pre-traind autoregressive LLM to produce satisfying results, without additional training or fine-tuning. The in-context learning samples are usually specifically designed for designated downstream task and can serve as auxiliary parameters attached to the model to guide the generation process. \citet{icl1-liu2021makes} explored the way to better design the context sample through distance metrics, and impact of different kinds of the distance metrics like Euclidean distance and so on. \citet{icl5-levy2022diverse} proposed a method to reinforce generalization ability through sample diversification. \citet{icl4-chung2024scaling} introduced a perplexity based method for designing in-context samples. \citet{icl2-sorensen2022information} conducted cross-lingual contextal learning experiments using clustering methods. \citet{icl3-tanwar2023multilingual} used ICL techniques to perform the alignment task of different languages.

\subsection{Automatic Depression Detection}
Plenty of works have been dedicated to automating the process of depression detection by implementing methods like natural language processing, machine learning, multimodal model LLM. To start with, researchers used traditional methods to tackle with the issue. For instance, \citet{Abdurrahim2024MentalHP} introduced a Convolutional Neural Network (CNN) and Bidirectional Long Short-Term Memory (BiLSTM) based deep learning model to perform depression detecting tasks upon social media posts. \citet{Cai2023DepressionDO} and \citet{10037751} have posed time series based LSTM to detect suicide risk. Meanwhile, there are also some works (e.g. \citealp{Wang2024SADTIMEAS, Yao2024Depression, YING2024106182, DAI20211040}), combining feature engineering with machine learning algorithms and deep neural network methods for diagnosing mental disorder. Multimodal methods \cite{tsai-etal-2019-multimodal,wang2020joint,Hazarika2020MISAMA} have also been generally applied in depression detection task. \citet{Ali2024LeveragingAA} leveraged audio and text modalities to analyze sentiment and mental health, with a method of prompt engineering. \citet{Ye2024DepMambaPF} extracted features of audio and video respectively and used the technique of mamba to fuse them and made collaborative classification for depression detection. \citet{10.1145/3627673.3679797} proposed HiQuE, hierarchically modeled the question and answer series in the interview dialogues for depression detection. Furthermore, LLMs like BERT, LLaMA and GPT have also been implemented by many works due to its reasoning ability. \citet{Lan2024DepressionDO}, \citet{Shah2024AdvancingDD}, \citet{Kuzmin2024MentalDD} implemented some prevailing LLMs and got decent performance on social media based dataset. \citet{10354236} developed a chat based method using GPT for interactive depression. \citet{Yang2023MentaLLaMAIM} proposed MentaLLaMA, a fine tuned version of LLaMA-2, which concentrated on interpretability of issues upon mental disorder. 

To the best of our knowledge, interactive depression detection methods based on explicit theme learning have not yet been explored in the field of automated depression detection. Unlike existing related work, our model is the first to learn theme correlation and design an interactive strategy to incorporate clinical feedback for preference learning.

\begin{figure*}[ht]
\centering
\begin{center}
\centerline{\includegraphics[width=\textwidth]{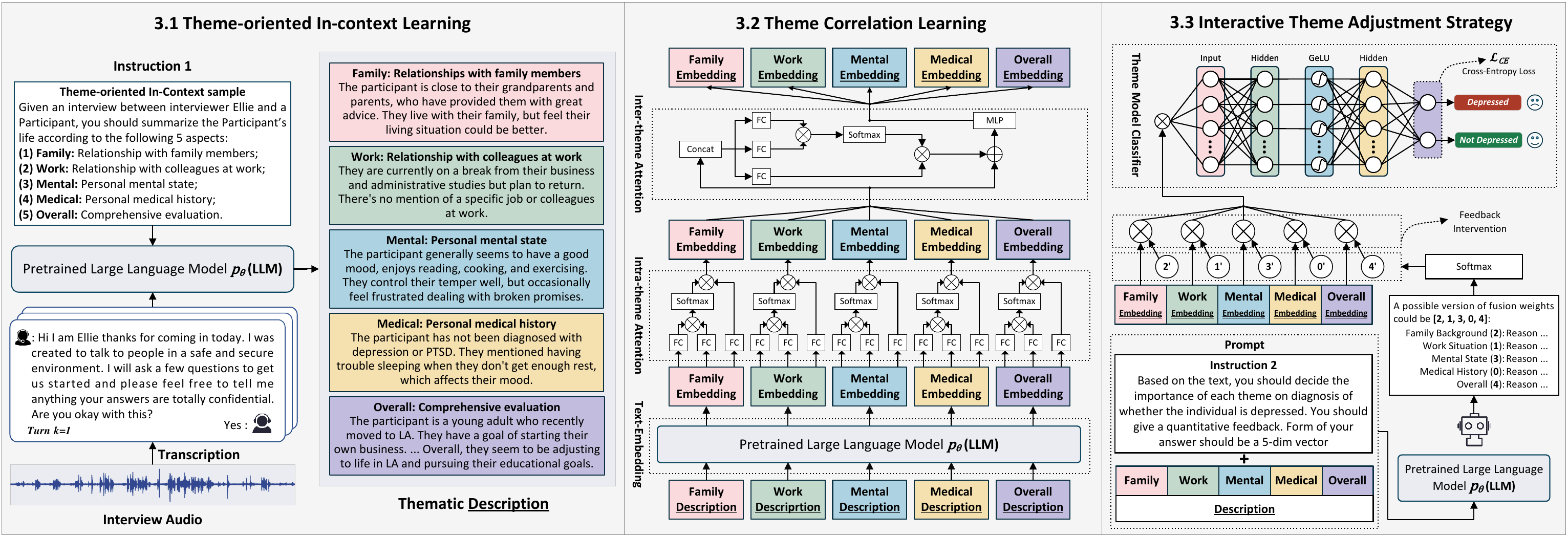}}
\caption{
Schematic illustration of the proposed PDIMC framework with three components. The theme-oriented in-context learning technique leverages the LLM to learn theme content from clinical interview dialogues. Theme correlation learning captures the inter- and intra-theme semantics related to depressive states. The interactive theme adjustment strategy utilizes the LLM to simulate clinical feedback, dynamically adjusting theme importance.
}
\label{fig:model}
\end{center}
\end{figure*}

\section{METHODOLOGY}
Our goal is to learn multiple themes $T_i,i\in{\mathcal{D}_{\{family,work,mental,medical,overall\}}}$ from multi-turn clinical interview dialogues $S$, then model both intra-theme and inter-theme correlation ($X^{inter}_i,X_i^{intra},i\in\mathcal{D}$). Additionally, we introduce feedback simulated by the LLM to imitate feedback of clinician, enabling preference learning to adjust themes. Finally, we fuse the adjusted themes $X_i^{fd},i\in\mathcal{D}$ to obtain final representation $X^{final}$ for depression prediction $\hat{y}$. In this section, we introduce each component of the proposed model, as illustrated in Figure \ref{fig:model}. Specifically, we first present the theme-oriented in-context learning module in Section \ref{TICL}, which extracts themes from clinical interview dialogues. Afterwards, we introduce the theme correlation learning module, which captures both intra-theme and inter-theme correlation in Section \ref{TCL}. Finally, we describe the interactive theme adjustment strategy in Section \ref{ITA}, which enables the model to focus on clinician-preferred themes for depression detection.

\subsection{Theme-oriented In-context Learning}
\label{TICL}
Clinical interview dialogues are structured around multiple themes, encompassing both thematic content and trivial details. The thematic content is closely related to the depressive state. To effectively capture these theme-related contents, we design a theme-oriented in-context learning module. This module leverages in-context learning techniques and LLM to extract depression-related themes while discarding irrelevant details. The in-context learning technique guides the LLM to generate output text $y_i$ based on the in-context template $I$ and user input sequence $X$. Formally, this process can be represented as:
\begin{gather}
    P(y_j|S,I) \triangleq p_\theta(X,I),\\
    \hat{y_j} = \textit{argmax}P(y_j|S,I),
\end{gather}
where $P$ represents the token probabilities and $\hat{y_i}$ denotes the token with the highest probability.  The operation of the theme-oriented in-context learning technique can be formally summarized as:
\begin{gather}
    T_{i} = p_\theta(S,I),i\in \mathcal{D},\\
    \mathcal{D} = \{{family}, {work}, {mental}, {medical}, {overall} \},
\end{gather}
where the LLM $p$ utilizes in-context prompt $I$ and model parameters $\theta$ to extract theme content $T$ from the clinical interview dialogues, filtering out trivial details and preserving information relevant to the depressive state. The in-context template, as illustrated in Figure \ref{fig:model}, consists of theme content (family, work, mental, medical, and overall) along with system prompts. 

\subsection{Theme Correlation Learning}
\label{TCL}
To fully leverage the advantages of themes, we model both intra-theme and inter-theme correlation. The intra-theme correlation aims to capture how key tokens within a theme influence the depressive state, while the inter-theme correlation focuses on learning how semantic relationships across themes contribute to depression assessment.  This module enhances the model's ability to highlight key semantics within a theme while also capturing semantic dependencies across themes. Figure \ref{fig:model} illustrates the proposed learning framework, demonstrating how these correlations contribute to a more comprehensive depression detection process. We employ the attention mechanism to learn both intra-theme and inter-theme correlation. Formally,
\begin{gather}
    X^{*} = f^{corr}_\phi(X),*\in{\{intra,inter}\},
\end{gather}
where $f^{corr}(\cdot)$ represents the correlation function with parameter $\phi$, while $X$ and $X^{*}$ denote the input before correlation and the output after correlation, respectively.  We first calculate the correlation matrix $A$. The intra-theme correlation matrix is computed by calculating the token similarity scores, while the inter-theme correlation matrix is determined by calculating the semantic similarity across themes. Formally,
\begin{equation}
        A(X) = softmax(\frac{X\cdot W_{Q} \cdot W_{K}^{T}\cdot X^{T}}{\sqrt{d}}),
\end{equation}
The computational operation of the correlation function $f^{corr}(\cdot)$ is as follows,
\begin{gather}
    f^{corr}_{\phi}(X) = A(X)\cdot X\cdot W_{V},
\end{gather}
where $A(\cdot)$ represents the correlation matrix function, while $W_{\{Q,K,V\}}$ denotes the learnable parameter matrix of $\phi$. We first use the pre-trained large language model $p_\theta$ to extract the features $X_i$ of the $i$-th theme $T_i$. Formally,
\begin{equation}
        X_{i} = p_\theta(T_{i})\in\mathbb{R}^{L_i\times d},i\in\mathcal{D},
\end{equation}
The form of intra-theme correlation learning $X_i^{intra}$ is as follows,
\begin{gather}
    X_i^{inter} = f^{corr}_{\phi_{inter}}(X_{i})\in \mathbb{R}^{L_i\times d},
\end{gather}
where $L_i$ represents the sequence length of the $i$-th theme, and $d$ denotes the feature dimension. The $\phi_{inter}$ denotes the learnable parameter matrix. After learning the token correlation semantics within the theme, we concatenate multiple themes along the sequence dimension. Formally,
\begin{equation}
        X^{inter} = Concat\{X_i^{inter}\}_{i=1}^{|\mathcal{D}|}\in\mathbb{R}^{(\sum L_i)\times d},
\end{equation}
immediately, we utilize the correlation learning function $f_\phi(\cdot)$ to learn the semantic of inter-theme correlation. Formally,
\begin{gather}
    X^{intra} = f^{corr}_{\phi_{intra}}(X^{inter})\in\mathbb{R}^{(\sum L_i)\times d},
\end{gather}
where $\phi_{intra}$ denotes the learnable parameter matrix. The inter-theme and intra-theme respectively highlight the importance of tokens within a theme and the importance of the theme itself.

\subsection{Interactive Theme Adjustment Strategy}
\label{ITA}
To address the issue that depression detection models cannot incorporate feedback from clinician, we have designed a customized interactive theme adjustment strategy. This strategy introduces simulated clinician feedback into the depression detection model, enabling the model to focus on the parts of interest to clinician. When incorporating external feedback information containing categories $c$ to constrain the model output in a generative model \cite{dhariwal2021diffusion,sanchezstay}, the model probability likelihood $\hat{\textbf{P}}_\Theta$ consists of the model prediction $\textbf{P}_\Theta$ and the prediction of the feedback information $\textbf{P}_\Phi$, resulting in the approximated modified distribution,
\begin{equation}
\hat{\textbf{P}}_\Theta(x|c)\propto \textbf{P}_\Theta(x)\cdot \textbf{P}_\Theta(c|x)^\gamma,     
\label{eq:cfg1}
\end{equation}
where $\gamma$ represents the constraint strength, which controls the model's degree of focus on the constraint. By removing the auxiliary classification task without classifier guidance using Bayes rule, the same model $\textbf{P}_\Theta$ simultaneously supports both conditional and unconditional predictions to reformulate Equation (\ref{eq:cfg1}) as: $\textbf{P}_\Theta(c|x)\propto\dfrac{\textbf{P}_\Theta(x|c)}{\textbf{P}_\Theta(x)}$. The sampling process of classifier-free guidance can be reformulated as:
\begin{gather}
    \hat{\textbf{P}}_\Theta(x|c)\propto\dfrac{\textbf{P}_\Theta(x|c)^\gamma}{\textbf{P}_\Theta(x)^{\gamma-1}},
\label{eq:cfg2}
\end{gather}
Taking the logarithm of both sides of Equation (\ref{eq:cfg2}) results in the following form,
\begin{gather}
            \text{log}\hat{\textbf{P}}(x|c) = \gamma\text{log}\textbf{P}_{\Theta}(x|c)-(\gamma-1)\text{log}\textbf{P}_\Theta(x)
        \label{eq:cfg3},
\end{gather}
let $\mathbb{P}(x) = (1-\gamma)\text{log}\hat{\textbf{P}}_\Theta(x)$, $\beta=\dfrac{\gamma}{1-\gamma}$, the probability distribution of the model's prediction can be rewritten as follows,
\begin{gather}
            \mathbb{P}(x) = \underset{\text{feedback\, prediction}}{\underline{\beta\text{log}\textbf{P}_\Theta(x|c)}}+\underset{\text{vanilla\,prediction}}{\underline{\text{log}\textbf{P}_\Theta(x)}},
\end{gather}
when introducing conditional constraints into the model prediction, the model's prediction is composed of the superposition of the prediction from the vanilla information and the prediction from the conditional constraint information. 

Based on the above theory, we integrate the classifier-free feedback of clinician on the theme with the vanilla theme information to diagnose depression. Formally,
\begin{gather}
    X_i^{fd} = X_i^{intra} + w_i^{fd}X_i^{intra},X_i^{fd}\in\mathbb{R}^{L_i\times d},
\end{gather}
The integration results for each theme are as follows:
\begin{equation}
        X_{final} = \sum_{i=1}^{|\mathcal{D}|}X_i^{fd},i\in\mathcal{D},w_i^{fd}\in{W^{fd}},
\end{equation}
where $w_{fd}$ represents the weight of the simulated clinical feedback from the LLM. Finally, we use the $Softmax$ function to normalize the weight of ($1+W^{fd}$) and integrate the weights of multiple themes for depression detection.

\subsection{Depression Detection Layer}
The final fused representation $X_{final}$ is passed through several fully connected layers and activation functions to obtain the prediction result $\hat{y}$. We compute the cross-entropy loss between the predicted value $\hat{y}$ and the ground truth $y$ to optimize the model:
\begin{equation}\mathcal{L}_{CE}=-\sum_i^{|N|}\mathbf{y}_i\log(\hat{\mathbf{y}}_i)+(1-\mathbf{y}_i)\log(1-\hat{\mathbf{y}}_i)
\end{equation}
where $N$ denotes the number of training samples.

\section{Experiments}
\subsection{Experimental Settings}
\paragraph{Datasets and Evaluation Metrics.}DAIC-WOZ \cite{daic-gratch2014distress} is a clinical interview dialogue dataset which is collected and released by the University of Southern California to help veterans back to civilian life. For metrics, we have conducted a series of experiments along with previous state-of-the-art hierarchical dialogue models \cite{10.1145/3627673.3679797} upon the indices of Accuracy, Precision, Recall, F1-score and G-means, and we also report the Precision, Recall, F1-score of Weighted Average version (denoted as WA$^*$Prec., WA$^*$Rec., WA$^*$F1 in Table \ref{tab:main}). Meanwhile, to compare against methods of latent theme, we reported F1 score of depression and non-depression on test set, following the previous work \cite{rinaldi2020predicting} (shown in Table \ref{tab:theme}). 

\begin{table*}[t]
\begin{center}
\setlength{\tabcolsep}{1.0mm}
\begin{tabular}{lcccccccc}
\hline
\textbf{Method} & \textbf{Accuracy} & \textbf{Precision} & \textbf{Recall} & \textbf{F1-Score} & \textbf{WA$^*$Prec.} & \textbf{WA$^*$Rec.} & \textbf{WA$^*$F1.}  & \textbf{G-Mean} \\
\hline
TFN & 0.81 & 0.67 & 0.72 & 0.68 & 0.84 & 0.78 & 0.81 & 0.699 \\
BiLSTM-1DCNN & 0.78 & 0.65 & 0.61 & 0.62 & 0.77 & 0.71 & 0.73 & 0.630 \\
MulT & 0.84 & 0.73 & 0.74 & 0.74 & 0.81 & 0.77 & 0.77 & 0.735 \\
MISA & 0.85 & 0.74 & 0.77 & 0.74 & 0.86 & 0.77 & 0.79 & 0.755 \\
D-vlog & 0.84 & 0.73 & 0.72 & 0.73 & 0.82 & 0.76 & 0.77 & 0.725 \\
BC-LSTM & 0.76 & 0.59 & 0.60 & 0.59 & 0.77 & 0.69 & 0.72 & 0.595 \\
EMSDL & 0.80 & 0.65 & 0.69 & 0.66 & 0.69 & 0.70 & 0.71 & 0.670 \\
ATSM & 0.81 & 0.67 & 0.71 & 0.70 & 0.85 & 0.73 & 0.77 & 0.690 \\
TopicModel & 0.78 & 0.63 & 0.60 & 0.62 & 0.81 & 0.71 & 0.74 & 0.615 \\
CADL & 0.83 & 0.71 & 0.71 & 0.71 & 0.85 & 0.73 & 0.77 & 0.710 \\
Speechformer & 0.83 & 0.70 & 0.72 & 0.70 & 0.78 & 0.76 & 0.76 & 0.710 \\
GRU/BiLSTM & 0.86 & 0.75 & 0.78 & 0.75 & 0.86 & 0.77 & 0.80 & 0.765 \\
HiQuE & 0.87 & 0.78 & 0.80 & 0.79 & 0.85 & 0.80 & 0.82 & 0.790 \\
\hline
\textbf{PDIMC} & \textbf{0.94} & \textbf{0.89} & \textbf{0.92} & \textbf{0.90} & \textbf{0.92} & \textbf{0.91} & \textbf{0.92} & \textbf{0.905} \\
\hline
\end{tabular}
\end{center}
\caption{
Performance comparison between our proposed PDIMC and several baselines on the DAIC-WOZ dataset.
}
\label{tab:main}
\end{table*}
\begin{table}[]
    \centering
    \setlength{\tabcolsep}{0.8mm}
    \begin{tabular}{cccc}
    \hline
    \textbf{Category} & \textbf{Model} & \textbf{F1 dep.} & \textbf{F1 non-dep.} \\
    \hline
    \multirow{5}{*}{\makecell[c]{Latent Theme \\ (Text)}}
    &PR & 0.45 & 0.82 \\
    &BERT & 0.44 & 0.84 \\
    &JLPC & 0.53 & 0.82 \\
    &JLPCPost & 0.52 & 0.85 \\
    \hline
     Explicit Theme & \textbf{PDIMC} & \textbf{0.87} & \textbf{0.92}\\
    \hline
    \end{tabular}
    \caption{Performance comparison between our proposed explicit theme model and implicit theme models.}
    \label{tab:theme}
\end{table}

\paragraph{Implementation Details.} We implemented PDIMC on Nvidia A100 GPU. Applying PyTorch library, our Adam Optimizer is configured with batch size of 32, learning rate of 10$^{-5}$, and training epochs of 80. For pretrained large language model, we employed Qwen2.5, an open source LLM, to extract text feature.

\subsection{Performance Comparison}
To demonstrate the effectiveness of our poposed PDIMC, we compared our proposed methods with several baselines. These methods are hierarchical dialogue models and latent theme models respectively. The hierarchical dialogue models includes: TFN \cite{zadeh-etal-2017-tensor}, BiLSTM-1DCNN \cite{Lin2020TowardsAD}, MulT \cite{tsai-etal-2019-multimodal}, MISA \cite{Hazarika2020MISAMA}, Depression Vlog (D-vlog) \cite{yoon2022d}, BC-LSTM \cite{poria2017context}, ERSDL \cite{satt2017efficient}, ATSM \cite{al2018detecting},  TopicModel \cite{gong2017topic},  CADL \cite{lam2019context}, Speechformer \cite{chen2022speechformer}, GRU/BiLSTM \cite{shen2022automatic}, Hierarchical Question Embedding Network (HiQuE) \cite{10.1145/3627673.3679797}.The latent theme models include: PR, BERT, JLPC, JLPCPost (all reported in \citealp{rinaldi2020predicting}).

The comparison results are summarized in Table \ref{tab:main} and \ref{tab:theme}. By comparing the results, we could draw the following conclusions. 
1) Compared to hierarchical dialogue models, our proposed method PDIMC shows remarkable improvement on all metrics. Among these baseline models, HiQuE \cite{10.1145/3627673.3679797} achieved the best performance in nearly every metric, except weighted average precision (WA$^*$Prec.). For macro metrics, our proposed method achieved 11\% better in Precision and F1-Score, 12\% better in Recall, compared to HiQuE. 
2) Compared to model of latent theme JLPC\cite{rinaldi2020predicting}, our model PDIMC of explicit theme achieved a 35\% improvement in the F1 score of depression. The experimental results indicate that explicit themes provide more clues related to depressive states than implicit themes.
This result suggests that the hierarchical dialogue model struggles to learn information related to depressive states when modeling hundreds of clinical interview dialogues. It further underscores the importance of learning themes and incorporating clinical feedback.

\begin{table}[]
    \centering
    \setlength{\tabcolsep}{0.7mm}
    \begin{tabular}{lcccc}
    \hline
    \textbf{Model} & \textbf{Acc.} & \textbf{WA$^*$Prec.} & \textbf{WA$^*$Rec.} & \textbf{WA$^*$F1} \\
    \hline
    w/o family & 0.85 & 0.85 & 0.85 & 0.85\\
    w/o work & 0.87 & 0.87 & 0.87 & 0.87\\
    w/o mental & 0.81 & 0.81 & 0.81 & 0.81\\
    w/o medical & 0.85 & 0.85 & 0.85 & 0.85\\
    w/o overall & 0.81 & 0.81 & 0.81 & 0.81\\
    \hline
    w/o TCL & 0.78 & 0.77 & 0.77 & 0.77\\
    w/o ITAS & 0.81 & 0.81 & 0.81 & 0.81\\
    \hline
    \textbf{PDIMC} &\textbf{0.94} & \textbf{0.92} & \textbf{0.91} & \textbf{0.92}\\
    \hline
    \end{tabular}
    \caption{
Ablation study to investigate the effect of different themes and tailored module.
    }
    \label{tab:ablation}
\end{table}

\subsection{Ablation Study}
To assess the effectiveness of each module we proposed, we conducted a series of decremental ablation study, removing one component once. The results are displayed in Table \ref{tab:ablation}, and we are giving a specific explanation. 1) \textbf{Without Single Theme}. In this section we alternately removed each theme and use the rest four themes to perform prediction. According to the results, removing any theme would cause a performance drop, from 5\% drop on WA$^*$F1 of work theme to 11\% of mental theme and overall theme. This demonstrated the irreplaceability of each theme, as they can maximize the utilization of the input text and collaboratively predict depression predisposition. 2) \textbf{Without TCL Module}. We removed the Theme Correlation Learning module, which means self-attention mechanism is disabled. From the result, we can see that the performance suffers the most performance decline, about 15\% on WA$^*$F1. This manifested the validity of self-attention mechanism which learns correlation both intra-theme and inter-theme and reinforces the accuracy of model predictions. 3) \textbf{Without ITAS Module}. This variant removes the Interactive Theme Adjustment Strategy module, and this means no discrepancy of weights would be applied during the multi-theme fusion process. Evidently, the 11\% drop on WA$^*$F1 indicated the indispensability of considering the prioritization of each theme when fusing them together. In general, the complete version of our model PDIMC outperformed the others with module removed in depression detection tasks. This result verify the effectiveness and complementary of the three components.

\subsection{Theme-oriented In-context Learning Analysis}
\begin{figure}
    \centering
    \includegraphics[width=\columnwidth]{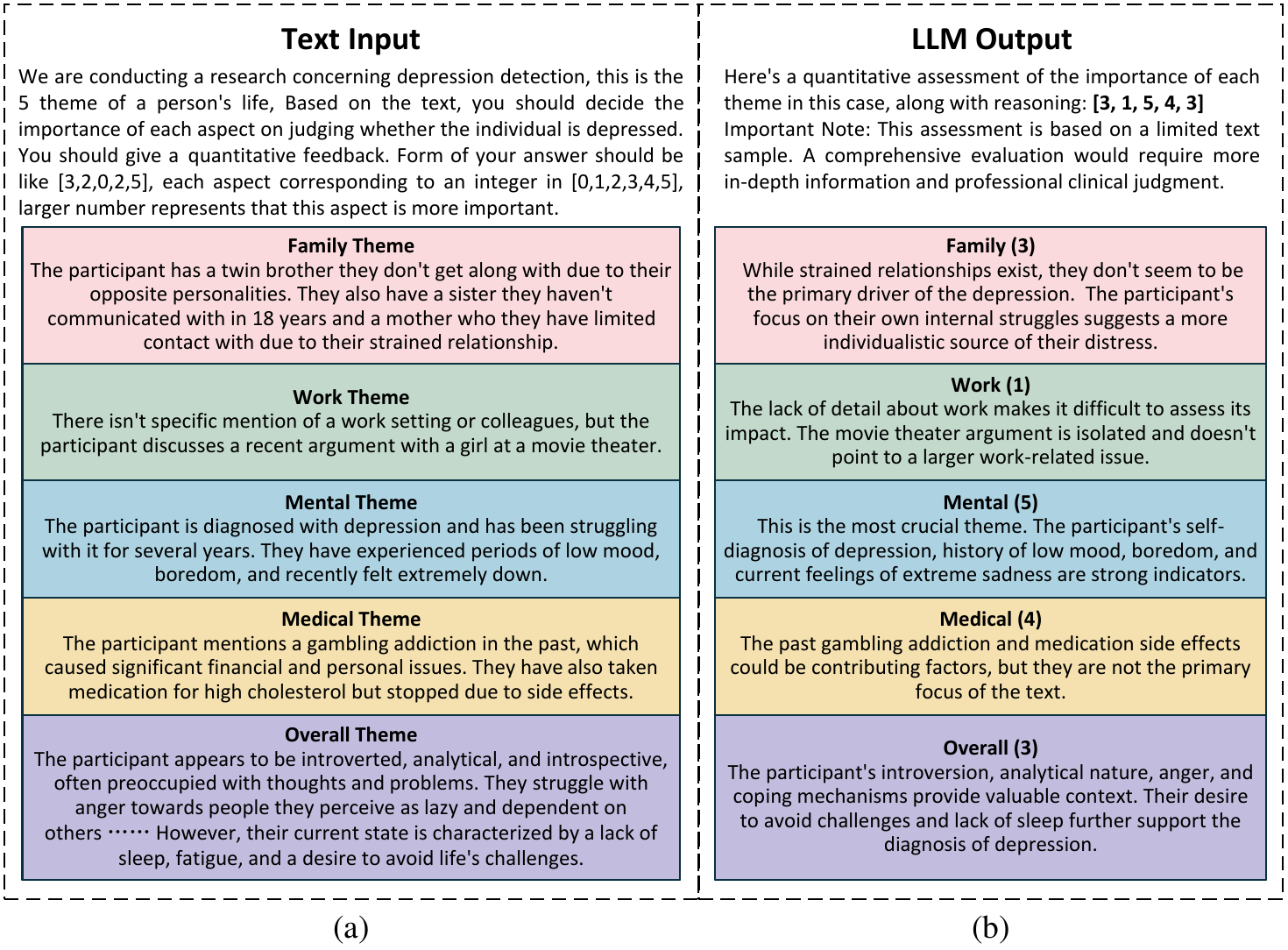}
    \caption{
     Visualization of the theme-oriented in-context learning technique illustrates the thems learned from clinical interview dialogues.
    }
    \label{fig:theana}
\end{figure}

Apart from achieving the superior performance, the key advantage of our model over other approaches is its ability to learn thematic content from complex clinical dialogues duo to theme-oriented in-context learning.
To this end, we carried out experiments to explore the ability of the module to extract thematic descriptions from the transcript text of the interview based on the in-context samples. Figure \ref{fig:theana} illustrates the themes extracted from hundreds of dialogue rounds, as well as the results of simulated clinical feedback generated by the large language model. By examining Figure 3, we can observe that the theme-oriented in-context learning technique accurately captures content related to family, work, mental health, and medical themes within the dialogue. Furthermore, the feedback simulated by LLM aligns well with common knowledge (for more details, refer to Section \ref{sec:itasa}). This result proved the capability of TICL module to extract valid thematic information from complex dialogues.

\subsection{Theme Correlation Learning Analysis}
\begin{figure}
    \centering
    \includegraphics[width=\columnwidth]{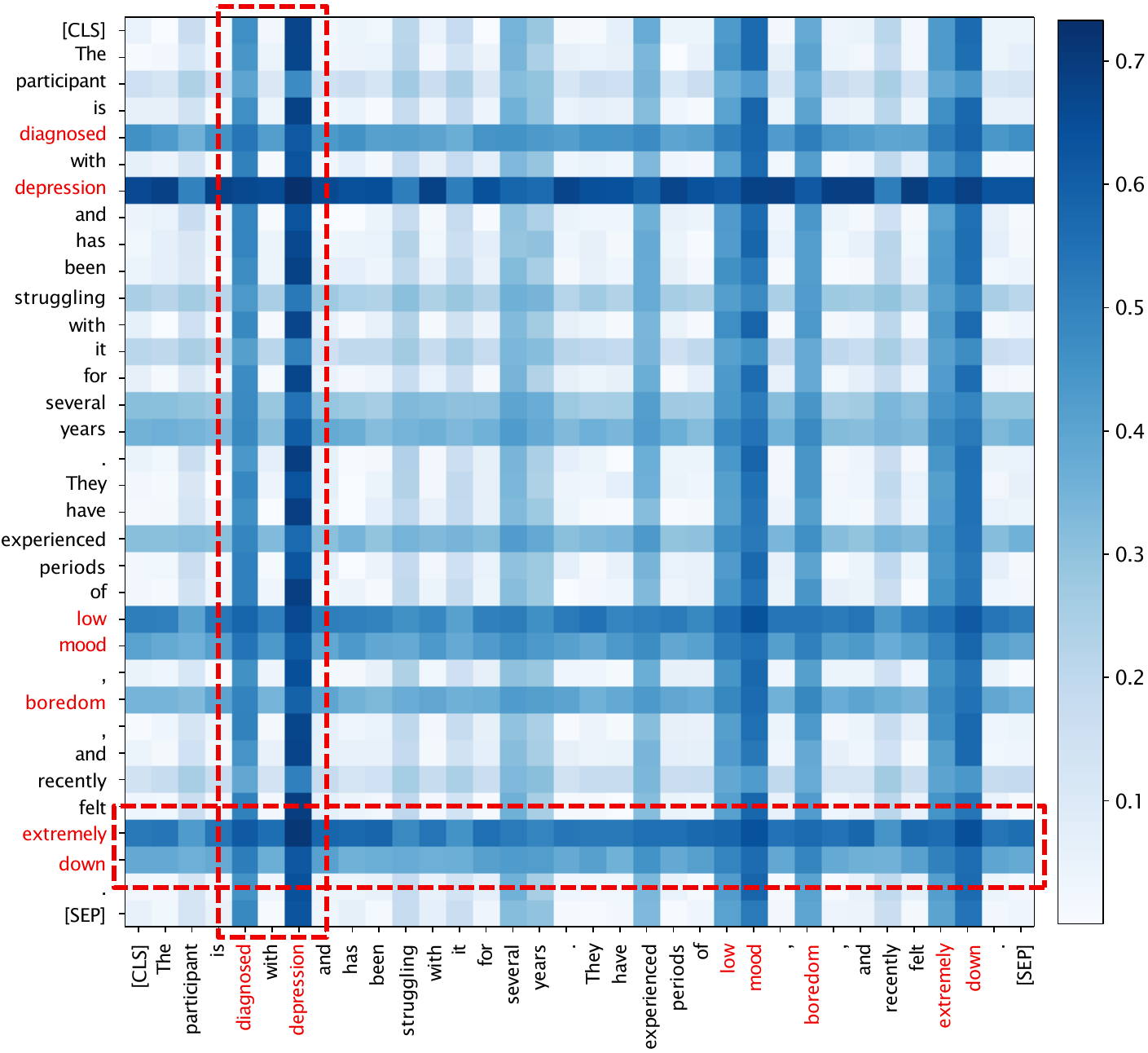}
    \caption{Visualization of intra-theme attention weights.}
    \label{img:TCL1}
\end{figure}
To qualitatively validate the effectiveness of the TCL module to capture correlation both intra-theme and inter-theme, we conducted experiments and visualized the map of attention distributions. Figure \ref{img:TCL1} displays the token weights of intra-theme correlation distribution. 
From Figure \ref{img:TCL1}, we could observe the token distribution within the "Mental" them and the corresponding attention weights. Some tokens, such as "diagnosed," "depression","extremely down" and "low mood," show a clear depressive tendency, and their corresponding attention weights are significantly higher than those of other tokens.
This result indicates that intra-theme correlation learning can highlight token semantic information related to depressive states, thereby enhancing the performance of the depression detection model.

\begin{figure}
    \centering
    \includegraphics[width=\columnwidth]{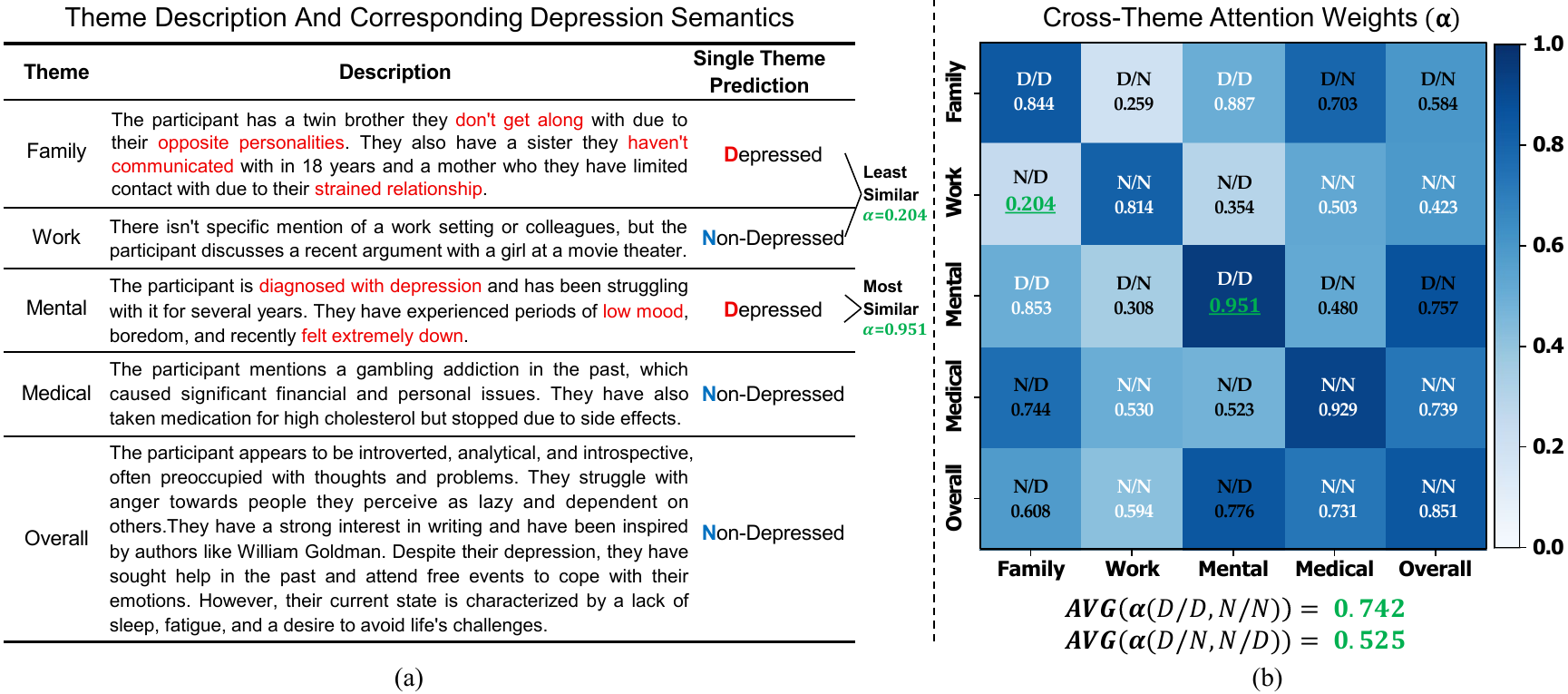}
    \caption{
    Visualization of inter-theme attention weights. (a) uses description of each single theme respectively to predict depression tendency, and we notate the predicted label upon the attention map of (b) (D for Depressed, N for Non-Depressed). Additionally, average attention scores of themes that share same and different predicted labels are calculated under the attention map in (b).
    }
    \label{img:TCL2}
\end{figure}

To gain the deep insights into our proposed inter-theme of theme correlation learning, we analyzed the weights of several inter-theme correlation. We obtained the following observations: 1) Themes with consistent depressive semantics have higher inter-theme correlation weights, while themes with inconsistent depressive semantics have lower correlation weights. For example, the highest similarity score among themes with consistent depressive semantics is 0.951, whereas the lowest similarity score among themes with inconsistent depressive semantics is 0.204. 2)The average weight of themes with consistent depressive semantics is higher than that of themes with inconsistent depressive semantics (0.742 vs. 0.525). 3)The inter-theme correlation weights are highly correlated with and accurately reflect depressive semantic information. These results indicate that inter-theme correlation learning effectively captures cross-theme semantic correlations, aiding the model's decision-making process and ultimately improving prediction accuracy.

\subsection{Interactive Theme Adjustment Strategy Analysis}
\label{sec:itasa}
\begin{figure}
    \centering
    \includegraphics[width=\columnwidth]{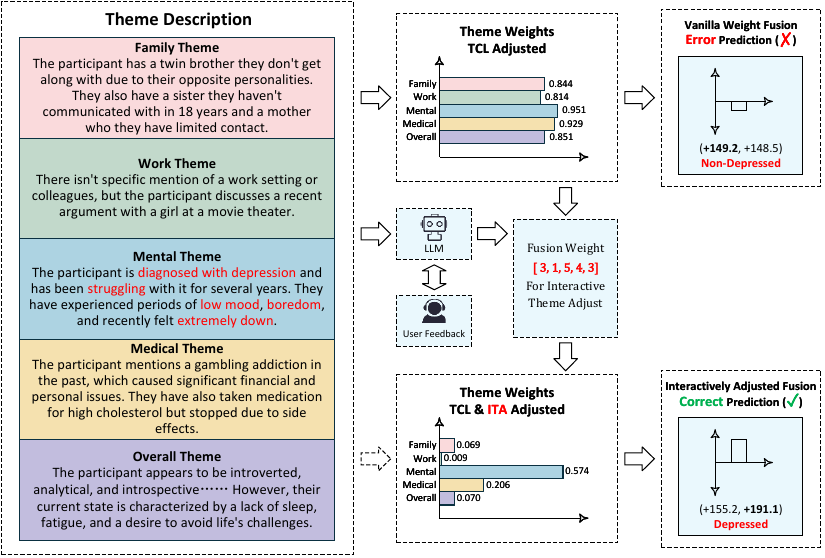}
    \caption{
    Visualization of the weights of each theme before and after applying the interactive theme adjustment strategy.
    }
    \label{fig:ITA}
\end{figure}

To validate the effectiveness of the interactive theme adjustment strategy, we compared the weights obtained through inter-theme correlation learning with those learned through the interactive theme adjustment strategy. Specifically, we first used the LLM to simulate clinical doctor feedback by scoring each theme, and then converted the scores into weights to intervene in the model's prediction. From Figure \ref{fig:ITA}, we can observe that the LLM’s scoring of theme importance is highly intuitive, such as the "Mental" theme receiving a high weight due to its direct correlation with depressive states. This result indicates that using the LLM to simulate clinician feedback is accurate.

From Figure \ref{fig:ITA}, we also could see that inter-theme correlation learning could, to some extent, capture semantic information related to depression, such as the higher weight of the "Mental" theme compared to other themes. However, the distinction is still not sufficiently clear, which may lead to incorrect predictions. After applying the interactive theme adjustment strategy, the weight of the "Mental" theme becomes significantly higher than that of other themes, leading to the correct prediction. This is because, during the multi-theme fusion phase, the distinction between themes is not clear enough, reducing the influence of key themes on the prediction results. These results demonstrate the importance of introducing simulated clinician feedback through the LLM, as it highlights key theme information to guide the model towards making accurate predictions.

\section{Conclusion}
In this paper, we proposed a novel interactive depression detection framework. The framework includes three components: 1) The theme-oriented in-context learning module learns interview theme-related information from multi-turn dialogues. 2) The theme correlation learning module mines correlation of inter-theme and intra-theme. 3) The interactive theme adjustment strategy introduces clinical feedback simulated by the LLM to guide the model in aligning with the preferences indicated by external feedback. Extensive experiments demonstrate the effectiveness and superiority of our proposed model. It not only achieves accurate depression diagnosis but also allows for clinical intervention, enabling integration of expert feedback into the diagnostic process.

\section*{Limitations}
There are two limitations in this study. First, although our approach achieves outstanding performance on depression detection datasets, it only utilizes textual data from clinical interview dialogues without incorporating multimodal information. Integrating multimodal data could potentially further enhance the model’s performance. Second, in designing the interactive depression detection framework, we use the large language model to simulate clinical feedback as external guidance. However, the professional reliability of this simulated feedback may require further evaluation.



\bibliography{custom}


%

\end{document}